\documentclass[conference]{IEEEtran}
\IEEEoverridecommandlockouts
\usepackage{amsmath,amssymb,amsfonts}
\usepackage{algorithmic}
\usepackage{graphicx}
\usepackage{textcomp, slashbox}
\usepackage{xcolor}

\ifCLASSOPTIONcompsoc
  \usepackage[nocompress]{cite}
\else
  \usepackage{cite}
\fi

\def\BibTeX{{\rm B\kern-.05em{\sc i\kern-.025em b}\kern-.08em
    T\kern-.1667em\lower.7ex\hbox{E}\kern-.125emX}}
    
\begin{document}

\title{Sentiment Analysis in Twitter Social Network Centered on Cryptocurrencies Using Machine Learning\\
}

\author{\IEEEauthorblockN{1\textsuperscript{st} Vahid Amiri}
\IEEEauthorblockA{\textit{Computer and Information Technology Department} \\
\textit{Razi University}\\
Kermanshah, Iran \\
vahid.kord68@gmail.com}
\and
\IEEEauthorblockN{2\textsuperscript{nd} Mahmood Ahmadi}
\IEEEauthorblockA{\textit{Computer and Information Technology Department} \\
\textit{Razi University}\\
Kermanshah, Iran \\
m.ahmadi@razi.ac.ir}

}

\maketitle

\begin{abstract}
The term cryptocurrency is an emerging topic in today's world, which has created a revolution in our vision in the field of investment and has caused changes in the world's financial systems. Cryptocurrency is a digital currency that uses blockchain technology with secure encryption. Due to the decentralization of these currencies, traditional monetary systems and the capital market of each they, can influence a society. Therefore, due to the importance of the issue, the need to understand public opinion and analyze people's opinions in this regard increases. To understand the opinions and views of people about different topics, you can take help from social networks because they are a rich source of opinions. The Twitter social network is one of the main platforms where users discuss various topics, therefore, in the shortest time and with the lowest cost, the opinion of the community can be measured on this social network. Twitter Sentiment Analysis (TSA) is a field that analyzes the sentiment expressed in tweets. Considering that most of TSA's research efforts on cryptocurrencies are focused on English language, the purpose of this paper is to investigate the opinions of Iranian users on the Twitter social network about cryptocurrencies and provide the best model for classifying tweets based on sentiment. In the case of automatic analysis of tweets, managers and officials in the field of economy can gain knowledge from the general public's point of view about this issue and use the information obtained in order to properly manage this phenomenon. For this purpose, in this paper, in order to build emotion classification models, natural language processing techniques such as bag of words (BOW) and FastText for text vectorization and classical machine learning algorithms including KNN, SVM and Adaboost learning methods Deep including LSTM and BERT model were used for classification, and finally BERT linguistic model had the best accuracy with 83.50
\end{abstract}

\IEEEpeerreviewmaketitle

\section{Introduction}
As the economic and social impact of cryptocurrencies is growing rapidly, the publication of related news and social media posts, especially tweets, is also increasing in this regard. It states whether they like other cryptocurrencies or not. Analyzing people's opinions about cryptocurrencies from different perspectives can help to identify public behavior towards the existence of cryptocurrencies and its importance in terms of people's opinions. On the other hand, analyzing all comments is an impossible task, even if it is possible, it is very difficult for humans to analyze these comments. Therefore, building a model for automatic analysis of people's opinions in this matter is of great importance. On the other hand, most of the research in the field of sentiment analysis is in English, and few works have considered the issue of sentiment analysis in Persian. While Persian language is spoken by millions of people all over the world and it is the official language of several countries like Iran, Afghanistan and Tajikistan. Therefore, the Persian language is a challenging language in the field of sentiment analysis, and few resources are available for processing the text of this language, while the processing of Persian language documents and texts has many challenges such as misspellings, spacing between words, finding roots and using It includes slang words\cite{ref1}, \cite{Ref11}.

 According to the previous works, it has been proven that the type of emotions is effective in predicting the price of cryptocurrencies. Cryptocurrency is a global currency, but until now, related work has been done on English tweets, and built models process English tweets \cite{ref2}. Considering that the number of Persian-speaking people is more than one hundred million people, building a model to classify the sentiments of Persian tweets alongside the models built for English tweets can usually give a more accurate output because it considers a larger statistical population.

In this paper, an attempt is made to build a model for classification of emotions (negative, positive and neutral) using natural language processing (NLP) techniques and machine learning methods on a set of Persian tweets collected about cryptocurrencies. Therefore, the effort is to use Persian language pre-processing techniques and text vectorization methods such as bag of words and FastText and BERT (Bidirectional Encoder Representations from Transformers) language models, as well as using machine learning algorithms and deep learning, it is possible to classify Persian tweets in the field of cryptocurrencies with acceptable accuracy, so that by using the built model, a statistical view of the level of people's interest in investing in this field can be obtained, and in the shortest time and with By spending the least amount of money, understand the emotional process of people in the society towards the field of cryptocurrencies and think of the necessary measures accordingly.

The rest of this paper is organized as follows. Section \ref{relatedworks} presents the related works on the sentiment analysis of cryptocurrency using using machine learning techniques. Section \ref{proposed} describes the our proposed method for sentiment analysis and classification. Section \ref{eval} describes the evaluation results and section \ref{conclusion} concludes the paper.

\section{Related works}
\label{relatedworks}
In this section, the related work the sentiment analysis of cryptocurrency using using machine learning techniques is presented.
In \cite{ref60}, a currency value fluctuation system based on users' sentiments was proposed, which predicted currency fluctuations by analyzing users' sentiments towards a specific currency. In \cite{ref61}, described the impact of social media on Bitcoin and show that social media platforms affect Bitcoin on an hourly basis. In addition, user comments and responses are also useful in determining the volatility of Bitcoin \cite{ref62}.

In \cite{ref63}, the ability of news and social media data to predict price volatility for three cryptocurrencies were analyzed. Daily news and social media data are labeled based on actual price changes one day in the future (one day and two days later) for each coin, rather than positive or negative sentiment. Using this approach, the model is able to predict price volatility directly instead of needing to predict sentiment. The final version of this model was able to correctly predict, on average, the days with the highest percentage increase and percentage decrease in the price of Bitcoin and Ethereum during 67 days including the test set.

In \cite{ref66}, a method to predict Bitcoin and Ethereum price changes using Twitter data and Google trends data was presented. By analyzing tweets, they found that tweet volume was a better predictor of price direction than tweet sentiment. Using a linear model that takes tweets and Google trend data as input, they were able to accurately predict the direction of price changes.

In \cite{ref71}, the relationship between Bitcoin attractiveness to investors and Bitcoin return, trading volume and perceived volatility on Twitter using linear and non-linear Granger causality tests was evaluated. The obtained results showed that the number of tweets significantly predicts the perceived volatility and the volume of Bitcoin future transactions.

In \cite{ref72}, a business intelligence model for predicting 5 high performing cryptocurrencies was presented. In this study, deep learning, linear regression, and support vector regression (SVR) are used to predict cryptocurrency prices. Emotions of some large events are also used to enhance the performance of these models.

In \cite{ref73}, they examine how well public sentiment on Twitter can be used to predict returns for some major cryptocurrencies. Using a sentiment analysis approach based on a specific cryptocurrency dictionary, financial data, and Granger causality test, it was found that Twitter sentiment has predictive power for the returns of cryptocurrencies.
This paper uses the several machine learning and deep learning models to analyze the Sentiment in Persian tweets.

\section{Proposed Method}
\label{proposed}
\begin{figure}[h!]
	\centering
	\includegraphics[scale=0.4]{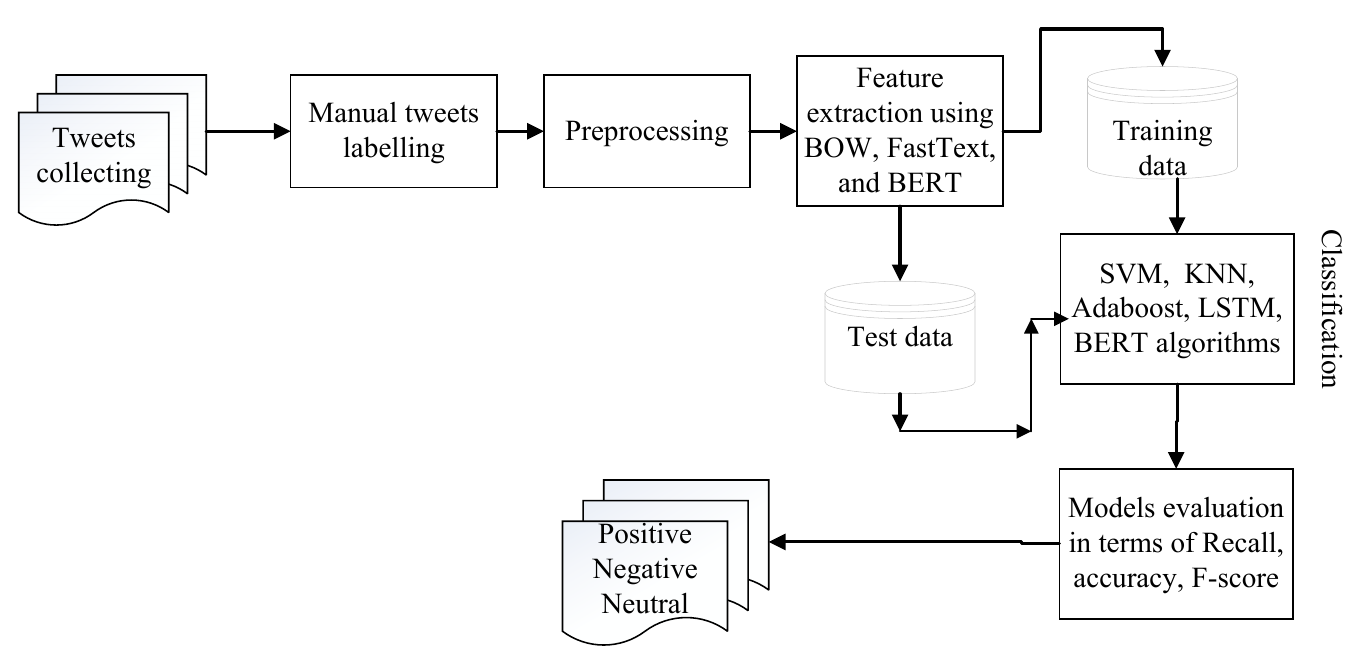}
	\caption{The block diagram of sentiment classification.}
	\label{fig1}
\end{figure}

In this section, the process of sentiment analysis of tweets and the methods used for modeling in this paper are explained. Fig \ref{fig1} depicts the block  diagram of sentiment classification system in this paper.

\subsection{The sentiment analysis process}
In order to perform twitter sentiment analysis (TSA) in this paper, the following steps have been implemented in order:
\begin{itemize}
\item Tweets collection.
\item Tweets labeling.
\item Tweets preprocessing.
\item Vectorization and feature extraction with Bow, fast text and BERT methods.
\item Emotion modeling and classification based on supervised learning, hybrid learning, deep learning and BERT (BERT) methods.
\item Evaluation of the built models based on accuracy criteria.
\end{itemize}
\subsubsection{ Collecting Tweets (Dataset Preparation)}

The first step of the sentiment analysis process involves collecting the required data. For this purpose, 4000 tweets have been collected in the form of an Excel file using the Twitter social network API and with the help of the search terms (in Persian language) \#crypto, \#crypto, \#digital currency, \#cryptocurrency and \#Bitcoin.

\subsubsection{Tweets labeling}
 One of the main challenges in creating new datasets for classification is data labeling. Manual labeling of emotions is done in two possible ways. The first method is through a group of experts and the second method is through crowdsourcing techniques. A crowdsourcing technique is the use of online platforms that allow anyone to manually tag tweets \cite{ref24}. In this paper, the first method was used to manually tag tweets into positive, negative and neutral classes.

\subsubsection{Tweets preprocessing}
Preprocessing of texts makes the data uniform and organized and ready for analysis and modeling. According to the research, only 33\% of the words of a text contain valuable information for extraction. Persian language processing is different from English language processing. Because in English all letters and all words are written separately and according to certain rules, while in Persian some letters are written together and some are written separately, some words are integrated and others are Interval or semi-interval is divided into two or more parts. Also, slang and colloquial words may be used in Persian texts (especially in user comments). Therefore, according to the mentioned cases, the pre-processing of Persian language texts has its own techniques. In the following, the pre-processing used on tweets in this paper is described.
\begin{itemize}
\item Remove punctuation marks: In the first step, punctuation marks such as exclamation mark (!), question mark (?), comma (,), period (.), quotation mark (:) and..., using regular expression method (RegEx) from tweets were deleted.

\item Removal of English words: Considering that the current research deals with the classification of the sentiments of Persian language tweets, all the words and characters that were in English were removed from the tweets. RegEx was also used for this.
\item  Delete numbers: All numbers were removed from the tweets. RegEx was also used for this.
\item Converting Arabic letters to Farsi: Using Persian and Arabic letters in the same words causes the computer to recognize them as different words, and this issue disrupts the performance of the model. For this reason, it is necessary to use only Persian letters. In this paper, the Persian library was used to convert Arabic letters into Persian. Persian library is a tool designed and implemented with the purpose of processing Persian texts. Among its capabilities, we can mention tokenization, normalization, root search, sentence search, stop words, etc. An example of Arabic letters to Farsi conversion is depicted in Fig. \ref{fig2}.
\begin{figure}[h!]
	\centering
	\includegraphics[scale=0.42]{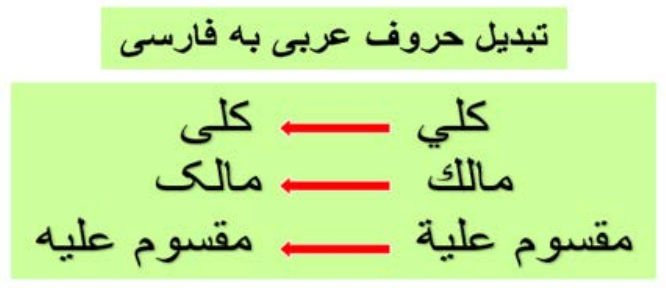}
	\caption{Arabic to Farsi conversion.}
	\label{fig2}
\end{figure}
\item  Remove stop words: Stop words are words that do not have much meaning and do not add special information to the sentence \cite{ref74}, but are used only to comply with the structure of the sentence and do not have much impact on analysis and modeling. For example, relative verbs (was, is, was, and...), conjunctions (to, and, from, to and...), pronouns (you, me, him, you, and...) and. .. can be named. To remove these words in this research, the list of stop words prepared by Kazem Taqwa et al. \cite{ref75} was used.

\item Normalization: Normalization is an operation that is used to unify texts and includes writing corrections, removing empty spaces, converting some spaces into half spaces, etc. Parsivar library was used for normalization in this research. Parsivar is a free and open source library based on NLTK. This library is provided by the University Jahad Information Technology Research Institute and has the necessary tools for pre-processing the Persian language.
\item Correct spelling mistakes: Due to the fact that tweets may contain spelling errors, correction of misspelled words is also done in the pre-processing steps. For this purpose, the Spell Check function from Parsivar library was used in this paper. An example of mistakes correction is depicted in Fig. \ref{fig3}
\begin{figure}[h!]
	\centering
	\includegraphics[scale=0.40]{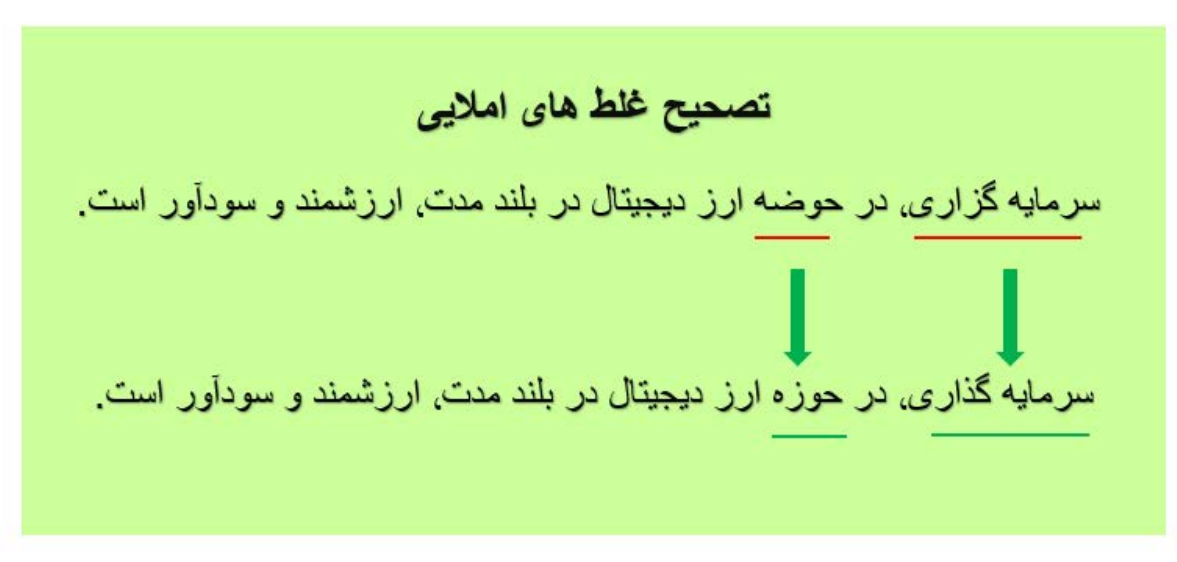}
	\caption{Mistake correction samples.}
	\label{fig3}
\end{figure}

\item Rooting (Stemming):
The last step of pre-processing in this research is rooting. Many words have more than one form that all these different forms have the same meaning despite the apparent difference. Finding word roots is one of the most important steps in text preprocessing. The purpose of rooting is to remove word stems (prefixes and suffixes) and determine the main root of the word, based on the rules of word formation. To find the roots of Persian words, a series of rules are used on these words. In this study, Parsivar library has been used for rooting.
\end{itemize}
\subsection{Wordcloud of Tweets}

\begin{figure}[h!]
	\centering
	\includegraphics[scale=0.25]{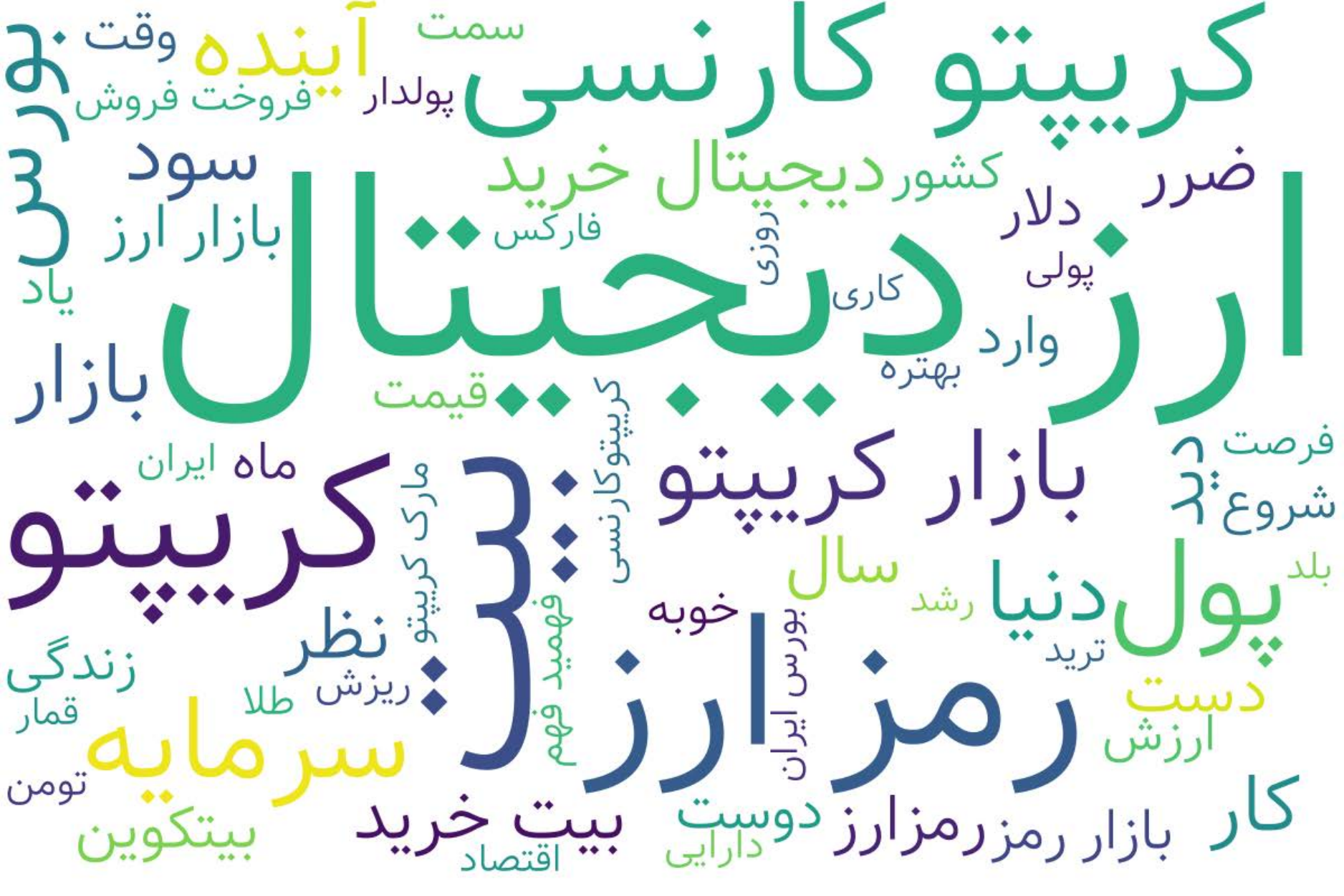}
	\caption{Word cloud of tweets.}
	\label{fig4}
\end{figure}

Cloud of words is a way to visualize textual data. It is actually a picture that shows us the words that are present in a text so that it changes its size according to the repetition rate of each word, i.e. how much a word is in the text. The more the input is repeated, the larger the final image will be. Figure \ref{fig4} shows the word cloud of the tweets of this research for quick, easy and meaningful analysis of the tweets.

\subsection{ Vectorization and feature extraction}
As mentioned, texts and documents in their original form are unstructured data. In text mining, before performing various operations e.g classification, it is necessary to convert these texts into structured and processable data for the computer. Therefore, it is necessary to identify and extract features by converting texts into vectors. For this purpose, there are different methods (based on frequency or based on word embedding), some of which are used in this paper, and are presented in the following.
\subsubsection{Bag of words (BOW)}
A bag of words is a representation of text that describes the occurrence of words in a document. It is called a "bag of words" because any information about the rules of the language, the order or structure of words, and the relationships between words in the document is left out. This model only considers whether known words (unique words in a body of text) are present in a document (regardless of where it is in the document). This method creates a matrix (fixed length vectors) where each unique word of the text corpus is represented by a column of the matrix and each document of the text corpus is a row in the matrix. Each value in the vector represents the number of repetitions of the word in the document. The bag-of-words model is commonly used in document classification methods where the (frequency) occurrence of each word is used as a feature to train a classifier \cite{ref76}.
\subsubsection{FastText}
The FastText language model is also a word embedding method that is an extension of the Word2Vec model. This method was proposed by Facebook in 2016. Most word representation learning methods, such as Word2Vec, ignore the morphology (internal structure) of words by assigning a separate vector to each word. This is a limitation, especially for languages with large vocabularies and rare words. Since many word formations follow rules, vector representations for morphologically rich languages can be improved using character-level information. FastText is a method based on the Skip-gram model, and each word w is represented as a bag of n-gram characters (a bag of sub-words). One of the features of FastText is that its pre-trained model is available for 157 world languages, including Persian, on the fasttext.cc website. In this research, the pre-trained model of the Persian language is used. It should be noted that in this research, the same arguments of the trained model are used, including the embedding size, which is considered equal to 300.
\subsection{ Classification}
In this paper, after vectorizing tweets and extracting features with the methods mentioned in the previous section, tweets were classified with machine learning methods, including supervised learning, hybrid learning, deep learning, and BERT's linguistic model, each of which is analyzed separately.
\subsubsection{BOW+SVM}
Scikit-Learn library was used to implement this model. The dataset was divided into two training and testing parts through the train\_test\_split function with a ratio of 80 to 20 and shuffled. CountVectorize function was used to implement the BOW vectorization method. OVR multi-class strategy and SVM classification algorithm were considered. The fit function was used for training and the predict function was used for testing.

Scikit-Learn library is one of the open source, useful, widely used and powerful libraries in the Python programming language, which is used for machine learning purposes and provides many practical tools for machine learning and statistical modeling of data such as classification, regression, clustering and It provides dimension reduction.
\subsubsection{ BOW+KNN}
In order to implement this model, the Python programming language library was used. First, using the train\_test\_split function, the dataset was divided into two parts of training and test data with a ratio of 80\% to 20\% in shuffle. Then, CountVectorize function was used to implement the BOW feature extraction method. Since the problem of this research is multi-class classification, the multi-class strategy of one class against the rest (OVR) along with the KNN classification algorithm from the mentioned library was considered to classify tweets into positive, negative and neutral classes. Using the fit function, the model was trained on the tweets of the training set, and the predict function was used to predict the tweets of the test set. It should be noted that the parameter k was considered the default value of 5.

\subsubsection{BOW+AdaBoost}

In this paper, in order to analyze emotions using the Adaboost algorithm, like the previous two methods, the Psychit Learning library was first used. With the help of train\_test\_split, the set of tweets was divided into two parts of training and testing with a ratio of 80\% to 20\% in shuffle. The CountVectorize function was used to implement the BOW vectorization method. In the following, the AdaBoost classification algorithm from the ensemble subset of the mentioned library was used as OVR to classify tweets. After training and testing with fit and predict functions, the accuracy of the model was calculated.

\subsubsection{FastText +LSTM }
First, the pre-trained language model specific to Farsi FastText was loaded using Gensim Python library, and then it was fine-tuned on the set of words of this research. The embedding size was considered equal to the value of the pre-trained model, i.e. 300. Subsequently, the sequential model of Cross Framework was used to build the model. In the sequential model, an embedding layer was first created and the weight matrix of tweets of the FastText model was placed in this layer. Then two layers of LSTM network with dropout=0.2 and finally two dense layers (the number of neurons in the last dense layer is 3) and activation functions relu and softmax respectively were created. Adam's optimization function and categorica\_crossentropy loss function and accuracy evaluation criterion were considered and then the model was trained with epoch number equal to 4 and bach\_size equal to 10. The ratio of dividing the dataset into training and testing was 80:20.
\subsubsection{BERT}
In this paper, the Python framework and the transformer library of the Python programming language were used to fine-tune the BERT model. Also, the trained Multilingual-BERT model was loaded and fine-tuned on the data of this research. In fine-tuning, the weights of the trained BERT model improve. A FFN was placed on the BERT output and the softmax activation function and the trained Multilingual-BERT model were used and fine-tuned on the sentiment analysis task. Here, 80\% of the data was assigned to the training data and 20\% of the data was assigned to the test data. Also, in this model, the number of iterations is 20, the batch size is 32, and the Adam optimization function is considered.
\section{ Evaluation}
\label{eval}
\subsection{Dataset}
\begin{figure}[h!]
	\centering
	\includegraphics[scale=0.9]{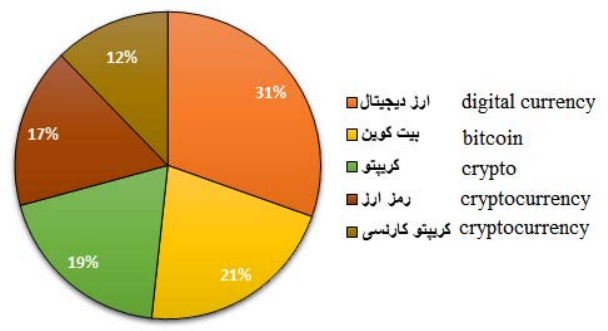}
	\caption{Frequency percentage of tweets based on search terms.}
	\label{fig6}
\end{figure}

The dataset of this research includes 4000 Persian tweets related to cryptocurrencies, which are considered to relate the collected tweets to words (in Persian) such as cryptocurrency, cryptocurrency, digital currency, crypto, and bitcoin. Figure \ref{fig6} shows the frequency of tweets in terms of the above words.

For sentiment analysis and classification of said tweets, three classes of positive sentiment, neutral sentiment and negative sentiment were considered. Tweets were tagged manually by It was found that 1958 tweets have a positive sentiment, 1597 tweets have a negative sentiment, and 445 tweets have a neutral sentiment.
\begin{figure}[h!]
	\centering
	\includegraphics[scale=0.9]{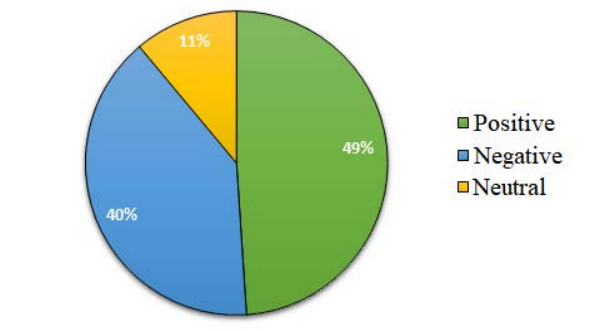}
	\caption{Frequency percentage of tweets assigned to positive, negative and neutral classes.}
	\label{fig7}
\end{figure}

 The system used in this paper has the following specifications.
\begin{table}[htb]
	\footnotesize
	\caption{System hardware specifications.}
	\centering
	\begin{tabular}{|c|c|}
		\hline
		Processor  & Intel Core i7    \\ \hline
		Graphic card    & Nvidia Geforce GT 540 2GB GDDR3  \\ \hline
		Main memory & DDR3 8GB 3200MHz        \\ \hline
		Storage         & Hard disk 1TB       \\ \hline
	\end{tabular}
	\label{table1}
\end{table} 

\subsection{Results}
\subsubsection{Classical machine learning algorithms}
First, the results obtained from classical algorithms based on machine learning are examined. It should be noted that in all the models below, 80\% of the data is assigned to the training data and 20\% to the test data.
\begin{itemize}
\item BOW+KNN model: In this method, feature extraction was done on the data using the BOW method, and then classification was done using the KNN algorithm. The accuracy obtained in this method was 74.50\%.
\item BOW+SVM model: In this model, after vectorization by BOW method, data classification was done with the help of SVM algorithm. The accuracy obtained in this method was 79.40\%.
\item BOW+AdaBoost model: Here, with the help of Adaboost algorithm, tweets were classified. It should be noted that the vectorization method and the ratio of dividing the dataset are the same as the previous two methods. The accuracy obtained was 77.50\%.
Tabel \ref{table2} shows the accuracy of these models.
\end{itemize}
\begin{table}[htb]
	\footnotesize
	\caption{Results of models based on machine learning.}
	\centering
	\begin{tabular}{|c|c|c|}
		\hline
		Algorithm  & Vectorization method & Accuracy  \\ \hline
		KNN    & Bag Of Words & 74.50\% \\ \hline
		SVM & Bag Of Words   &   79.40\%  \\ \hline
		AdaBoost         & Bag Of Words  & 77.50\%    \\ \hline
	\end{tabular}
	\label{table2}
\end{table} 
From these results, it can be seen that among the classic machine learning methods, BOW+SVM model performed better than other models and BOW+KNN had the weakest result.

\subsubsection{Algorithms based on deep learning}
In this section, the results related to the methods based on deep learning and BERT will be discussed.
\begin{itemize}
\item FastText+LSTM model: In this method, using the FastText language model, vectorization was done on the data, then the data was given to the combined LSTM algorithm to classify the data. The number of epochs in this method was equal to 4. The activation function of Relu and Softmax types and the optimizer function of Adam type were selected and categorical\_crossentropy as loss function was considered. Finally, this model obtained the accuracy of 82.32\%.
\item BERT model: In this method, the pre-trained BERT method was used in both vectorization and classification cases. The accuracy obtained by this model on the data of this research was 83.50\%. In the following, the results obtained from these models can be seen in the Table \ref{table3}.
\end{itemize}
\begin{table}[htb]
	\footnotesize
	\caption{Results of models based on deep learning.}
	\centering
	\begin{tabular}{|c|c|c|}
		\hline
		Algorithm  & Vectorization method & Accuracy  \\ \hline
		LSTM    & FastText & 74.50\% \\ \hline
		BERT & BERT   &   79.40\%  \\ \hline
			\end{tabular}
	\label{table3}
\end{table} 

Table \ref{table4} depicts the performance of the proposed method in terms of F1-score and recall metrics.

\begin{table}[htb]
	\footnotesize
	\caption{Results of models based on deep learning.}
	\centering
	\resizebox{0.5\textwidth}{!}{
	\begin{tabular}{|c|c|c|c|c|c|}
		\hline
	\backslashbox{Metric}{Model}	&Bow+KNN&Bow+SVM&Bow+Adaboost&FastText+LSTM& BERT \\ \hline
Recall	&65.7&64.2&63&69&73	 \\ \hline
F1-score	&72.10&74.5&72.5&77.25&78.3	  \\ \hline
			\end{tabular}}
	\label{table4}
\end{table} 

As can be seen from the results the performance of different machine learning and deep learning methods that were used to analyze the sentiments in Persian tweets related to the topic of cryptocurrency were investigated and determined using the evaluation criteria of Accuracy, Recall and F1-Score. Based on the results, the deep learning methods performed better than the classic machine learning algorithms and finally the proposed BERT model achieved the highest accuracy with 83.50\% accuracy compared to other used models. Therefore, according to the results stated in this section, the use of this model was effective for feature extraction and modeling, and according to the results obtained, it can be claimed that this model has a better ability than other models in solving the problem of classifying Persian tweets.

\section{Conclusion}
\label{conclusion}
As the economic and social impact of cryptocurrencies is growing rapidly, so is the spread of news and related posts on social media, especially Twitter. Investors' feelings are actually their behavior and general attitude towards financial markets such as cryptocurrencies. The topics discussed by these people express the importance of cryptocurrencies and the public opinions of people in this field and whether people in general have a desire to invest in cryptocurrencies or not. Analyzing people's opinions about cryptocurrencies from different perspectives can help to identify the public's position towards cryptocurrencies. On the other hand, analyzing all opinions is an impossible task, even if it is possible, analyzing these opinions is very difficult and time-consuming for humans. Therefore, it is very important to build a model to automatically analyze the opinions of people in the community about cryptocurrencies. In this research, using natural language processing techniques, machine learning and deep learning, sentiment analysis was done on Persian tweets related to cryptocurrencies in order to build a model for automatic classification of sentiments (negative, positive and neutral). For this purpose, first about 4000 Persian tweets collected on the subject of cryptocurrencies were pre-processed, then emotion modeling was done with the help of machine learning and deep learning methods. In the end, the BERT method was the best among all models by recording 83.50\% accuracy, and then FastText + LSTM model was ranked next with 82.32\% accuracy. One of the important points in this research was that methods based on deep learning performed better than classical algorithms based on machine learning. Therefore, it can be claimed that these algorithms appear stronger in the task of sentiment classification in Persian tweets with the topic of cryptocurrencies.



\bibliographystyle{IEEEtran}
\bibliography{ref}

\end{document}